\documentclass[letterpaper, 10 pt, conference]{ieeeconf}  

\IEEEoverridecommandlockouts                              

\usepackage{xcolor}
\usepackage{graphicx}
\usepackage{subfigure}
\usepackage{amsmath} 
\usepackage[ruled,vlined]{algorithm2e}
\usepackage{tikz}
\usetikzlibrary{matrix,shapes,arrows,positioning,chains,shapes.geometric,
				fit,decorations.markings,positioning}
\usepackage{listings}
\usepackage{cite}
\usepackage[colorlinks=true, 
			pdfstartview=FitV, 
			linkcolor=black, 
			citecolor=black, 
			plainpages=false, 
			pdfpagelabels=false, 
			urlcolor=black]{hyperref}
\usepackage[all]{hypcap}

\usepackage{import}
\usepackage{acro}

\acsetup{first-style=short}

\DeclareAcronym{cdt}{
    short = Constrained Delaunay Triangulation  ,
    long  = Constrained Delaunay Triangulation ,
    class = abbrev
  }
\DeclareAcronym{ddt}{
short = Dynamic Delaunay Triangulation ,
long  = Dynamic Delaunay Triangulation ,
class = abbrev
}
\DeclareAcronym{dt}{
  short = Delaunay Triangulation ,
  long  = Delaunay Triangulation ,
  class = abbrev
}
\DeclareAcronym{tm}{
  short = triangulation mesh ,
  long  = triangulation mesh ,
  class = abbrev
}

\DeclareAcronym{anchor}{
  short = $tri_{anchor}$ ,
  long  = The number of angels per unit area ,
  sort  = a ,
  class = nomencl
}

\DeclareAcronym{tevent}{
  short = $\tau$ ,
  long  = The number of angels per unit area ,
  sort  = a ,
  class = nomencl
}

\DeclareAcronym{teta}{
  short = $t_{eta}$ ,
  long  = The number of angels per unit area ,
  sort  = a ,
  class = nomencl
}

\DeclareAcronym{channel}{
  short = $c_t$ ,
  long  = channel found by graph search ,
  sort  = a ,
  class = nomencl
}

\DeclareAcronym{channelseg}{
  short = $c_i'$ ,
  long  = channel found by graph search ,
  sort  = a ,
  class = nomencl
}

\DeclareAcronym{finalchannel}{
  short = $C$ ,
  long  = channel found by graph search ,
  sort  = a ,
  class = nomencl
}

\title{\LARGE \bf
Spatial Constraint Generation for Motion Planning in Dynamic Environments
}

\author{Han Hu$^{1*}$ and Peyman Yadmellat$^{2}$
\thanks{$^{1}$Department of Mechanical \& Industrial Engineering, University of Toronto, Toronto, Ontario, Canada M5S 3G8}
\thanks{$^{2}$Noah’s Ark Lab., Huawei Technologies Canada, Markham, Ontario,
Canada L3R 5A4}
\thanks{$^{*}$The work was done during the author's internship at Noah's Ark Lab., Huawei Technologies Canada.}}

\begin{document}

\maketitle
\thispagestyle{empty}
\pagestyle{empty}

\begin{abstract}
This paper presents a novel method to generate spatial constraints for motion planning in dynamic environments. Motion planning methods for autonomous driving and mobile robots typically need to rely on the spatial constraints imposed by a map-based global planner to generate a collision-free trajectory. These methods may fail without an offline map or where the map is invalid due to dynamic changes in the environment such as road obstruction, construction, and traffic congestion. To address this problem, triangulation-based methods can be used to obtain a spatial constraint. However, the existing methods fall short when dealing with dynamic environments and may lead the motion planner to an unrecoverable state. In this paper, we propose a new method to generate a sequence of channels across different triangulation mesh topologies to serve as the spatial constraints. This can be applied to motion planning of autonomous vehicles or robots in cluttered, unstructured environments. The proposed method is evaluated and compared with other triangulation-based methods in synthetic and complex scenarios collected from a real-world autonomous driving dataset. We have shown that the proposed method results in a more stable, long-term plan with a higher task completion rate, faster arrival time, a higher rate of successful plans, and fewer collisions compared to existing methods.
\end{abstract}

\section{INTRODUCTION}



Motion planning in a dynamic environment is a cornerstone of autonomous driving and a challenging engineering problem. Motion planning methods often require spatial constraints provided by semantic maps (\textit{e.g.}, centerline, lane boundary) to generate a trajectory. This requirement results in motion planning failure where the spatial constraint is absent or invalid; unmapped roads, unstructured open areas, congested traffic, partially obstructed single lane, etc.

To address this issue, we propose a new method to generate spatial constraints from the triangulation mesh of an environment that can be applied to autonomous driving. Existing triangulation-based path planning methods construct a triangulation mesh from nodes in the environment. The nodes can be the vertices of objects or center of objects. Subsequently, a channel which is a sequence of free triangles within the triangulation mesh, connecting the start location to the goal location, is found via a graph search method. The channel can serve as a spatial constraint for motion planning algorithms (e.g. Hybrid A* or Modified Funnel Algorithm). As the main advantage, the triangulation representation reduces the overall time complexity as it results in a substantially smaller adjacency graph compared to grid-based methods~\cite{kallmann2005path}.

Most known methods for motion planning inside a \ac{tm} (\textit{e.g.}~\cite{kallmann2005path, chen2010enhanced, yan2008path, kallmann2014navigation, perkins2013field, demyen2006efficient, broz2014dynamic}) ignore dynamic objects, assume a static environment, and rely on repeated replanning to handle dynamic objects. These methods typically replan through dynamically maintained triangulation structure~\cite{devillers1992fully, mostafavi2003delete, chew1989constrained, edelsbrunner1996incremental}, incremental, or anytime variant search algorithms~\cite{likhachev2005anytime,koenig2002d,ferguson2005field} for efficient replanning, or a combination of both. The primary challenge with these methods is that they are not flexible to changes in the environment. This is because the channel generation scheme in these methods often suffers from two fundamental limitations: a) invariant triangulation mesh connectivity assumption and b) masked dynamic nodes, which causes the channel. Combined, these two limitations can result in volatile and near-sighted path planning, which is not suitable for autonomous driving vehicles.
%

The first limitation is that the nodal connectivity of the triangulation mesh is assumed to be invariant over time. This is not a valid assumption as a subset of the channel no longer exist when the node connectivity changes. Motion planning based on a channel that does not exist can result in a near-sighted plan, leading to unfavorable situations such as path planner failures, sudden stops, and unrecoverable states. The second limitation is that the existing methods only consider the interaction between adjacent nodes. However, this approach causes the motion information of the non-adjacent dynamic nodes to be masked and ignored by the adjacent nodes' connectivity. As a result, the ego vehicle will have latent responses to dynamic nodes (e.g. pedestrians, vehicles).



To address the first limitation, we propose a novel, modular system to generate a sequence of connected channels segments. Each segment corresponds to a change in node connectivity at different points in time. This sequence of segments is generated by identifying the segment affected by the change and replacing it with a segment from the triangulation mesh connectivity after the change. We also propose a method to address the second limitation by transmitting distant nodes' motion information along the triangulation mesh edges. This is achieved by projecting each node's velocity vector to the adjacent nodes along the edges, thereby giving each node an estimated motion information of the distant nodes. Together, the proposed systems and method generate a sequence of channel segments. Each segment can be used by a motion planner as the spatial constraint to generate a partial trajectory, which is then combined into the overall plan, thereby finding a stable, long-term plan in a dynamic environment.

The main contributions of this paper are:
\begin{itemize}
	\item A spatial constraint generation system that predicts and plans in the triangulation mesh of the environment at different points in time.
	\item A method of transmitting motion information of dynamic nodes within the triangulation mesh to distant nodes using the triangulation mesh edges.
\end{itemize}


\section{RELATED WORKS}\label{sec:related_works}


	


Most existing triangulation mesh planning methods focus on fast re-planning. Kallmann~\cite{kallmann2005path} used A* in with a dynamically maintained Constraint Delaunay Triangulation~\cite{chew1989constrained} to update node position during runtime. 
Chen~\textit{et al.}~\cite{chen2010enhanced} enhanced Dynamic Delaunay triangulation (DDT)~\cite{devillers1992fully} with a heuristic based algorithm for path planning to handle nodes that were newly identified by sensors.
The Target Attraction Principle was proposed in~\cite{yan2008path} to position the nodes of a dual graph to a \ac{cdt} as a method to optimize the path cost.

Numerous anytime and incremental variants of A* search algorithms were reviewed in~\cite{kallmann2014navigation} path planning methods for a polygonal mesh. For instance, Anytime Repairing A* (ARA*)~\cite{likhachev2004ara} attempts to return the best path found for a given time budget by iteratively improving an initial suboptimal solution. Incremental search algorithms such as D* Lite~\cite{koenig2002d} reuses and repairs an initial solution found using A* at each re-planning cycle to adapt to the changes instead of re-planning from scratch. Field D*~\cite{ferguson2005field} was used in~\cite{perkins2013field} for efficient re-planning within triangulation and tetrahedral mesh structures. Demyen~\textit{et al.}~\cite{demyen2006efficient} introduced Triangulation Reduction A*, for path planning on a reduced dual graph of the environment's \ac{tm}. As an alternative to re-planning,~\cite{broz2014dynamic} proposed the Gaps Filling algorithm to repair the previously found path by reconnecting broken path segments using Dijkstra's algorithm. It also employed an incremental insertion method~\cite{edelsbrunner1996incremental} for efficient triangulation structure updates. To account for object dynamics, Timed A* was proposed in~\cite{cao2019dynamic} for path planning in the dual graph of the Delaunay triangulation (DT). It considers the object dynamics by modifying the A*’s cost and heuristic function to include the ego vehicle's estimated time of arrival.

The above-mentioned methods do not allow for the consideration of non-adjacent nodes, and do not account for the changes to the triangulation mesh connectivity during planning. Distinct from previous work, our proposed method generates a sequence of connected channel segments to account for dynamic objects and their effect on the triangulation mesh connectivity.

\section{METHODOLOGY}\label{sec:methodology}
\begin{figure}[tb]
	\centering
	\subfigure[]{%
		\includegraphics[trim={0 1.5cm 0 0}, clip, width=0.15\textwidth]{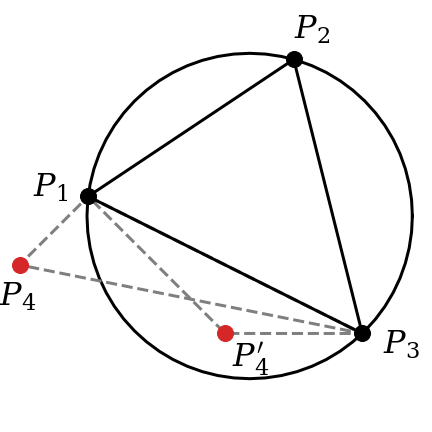}
		\label{fig:topological trigger}}
	\subfigure[]{%
		\includegraphics[trim={0 1.5cm 0 0}, clip, width=0.15\textwidth]{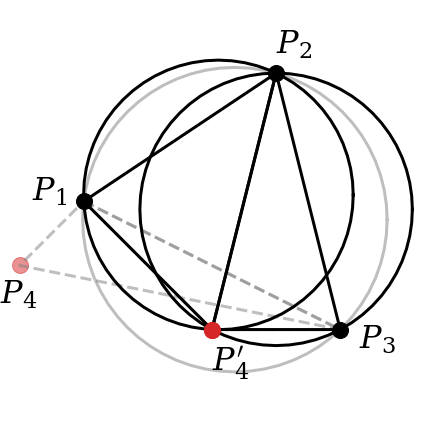}
		\label{fig:topological flip}}
		\vspace{-8pt}
	\caption{Topological Event: a) topological event triggered by $P_4$ moving to $P_4'$; and b) violation repaired with edge flipping.}
	\label{fig:topological event}
\end{figure}

We assume a point representation of the static and dynamic nodes extracted from the environment, where each node represents a vertex, a point on the edge, or the centroid of the object. Each node is described by a 5D vector $\{x, y, \dot{x}, \dot{y}, r\}$, where $r$ is the radius of the node. We also assume that all nodes follow a linear motion model. The commonly used Delaunay Triangulation is employed to generate a triangulation mesh from these nodes. A Delaunay Triangulation structure over a set of points P must satisfy the condition that no $p \in P$ lies inside a triangle's circumcircle. The event which violates this condition is known as a topological event shown in Figure~\ref{fig:topological event}(a). This violation can be locally repaired by updating the edge connectivity as shown in Figure~\ref{fig:topological event}(b). Repairing the Delaunay Triangulation structure replaces the two triangles that cause the violation with two new triangles. This changes the edge connectivity of the triangles, thereby introducing new node adjacency and dual graph nodes.

Our objective is to find a sequence of spatially and temporally connected channel segments $\{C : C = \bigcup_{t=0}^N c'_t\}$, where $\{c'_t : c'_t =  \bigcup_{j=0}^{k} \Delta_{t,j} | k\leq n\}$ is a subsequence of $c_t$, such that $\Delta_{t,j} \in c'_t$ is a subset of triangles $\Delta_{t,i} \in c_t$ that are unaffected by the changes in the edge connectivity of the triangulation mesh at time $\tau_t$. The channel ${\{c_t : c_t = \bigcup_{t=0}^n \Delta_{t,i}\}}$, is a sequence of free triangles $\Delta_{t,i}$ in the triangulation mesh connectivity at $\tau_t$, the triangle $\Delta_{t,0}$ contains the position of the ego vehicle and $\Delta_{t, n}$ contains the goal point. $c_t$ is found by applying graph search on the dual graph of the triangulation mesh.

We consider $\Delta_{t,i}$ to be unaffected by changes in the triangulation mesh connectivity if i) there are no change to the edge connectivity of $\Delta_{t,i}, i\in\{0,\dots,n\}$ or ii) the first edge connectivity change of $\Delta_{t,i}$ occurred when the ego vehicle is located in $\Delta_{t,e} \in c_t$, and $e>i$. The first condition limits our consideration to the triangles that makes up the channel, since the changes to the triangles outside of the channel does not change the edge connectivity of the channel. The second condition excludes any $\Delta_{t,i}$ that had a change in edge connectivity after the ego vehicle already past it as these changes do not affect the ego vehicle.

Following the second condition, the last triangle of $c'_t$, \textit{i.e.} $\Delta_{t,k}$, is determined by first finding the first triangle in $c_t$ that had a change in its edge connectivity, $\Delta_{t,m}$, that satisfies the following: ${\Delta_{t,m} = \Delta_{t,i} \in c_t | e \leq i}$. If $e<m$, then $k=e$. If $e = m$ then $k=m-1$, because we define $c'_t$ to be triangles that are unaffected by changes in their edge connectivity. We do not consider $e>m$ as it is excluded by the second condition. $\Delta_{t,e}$ is the triangle that contains the position of the ego vehicle estimated at time $\tau_{t+1}$, where $\tau_{t+1}$ is when the first edge connectivity change of $\Delta_{t,m}$ occurred.

Subsequent channel segment $c'_{t+1}$ is similarly found by identifying the segment of $c_{t+1}$ that is unaffected by changes in the edge connectivity of the triangulation mesh at time $\tau_{t+1}$. Where $c_{t+1}$ is from the expected triangulation mesh connectivity at $\tau_{t+1}$ under a linear motion model.

We ensure $c'_t$ and $c'_{t+1}$ is spatially and temporally connected by setting the last triangle of $c'_t$ to be the first triangle of $c'_{t+1}$ at $\tau_{t+1}$, i.e. $\Delta_{t,k} = \Delta_{t+1,0}$.

\begin{figure}[tb]
	\centering
	\includegraphics[width=0.35\textwidth]{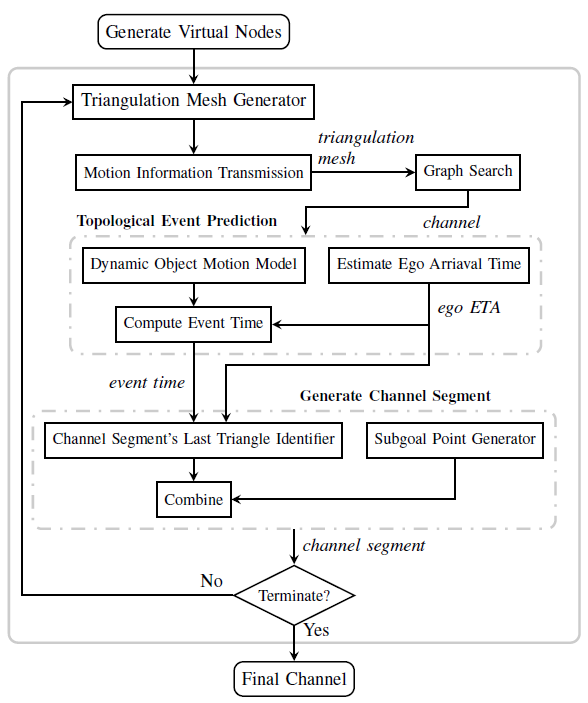}
    \vspace{-6pt}
	\caption{Process diagram of the proposed method.}
	\label{fig:process diagram}
\end{figure}

The following sections describe the main modules of the system shown in Figure~\ref{fig:process diagram} in detail.

\subsection{Generate Virtual Nodes}
For autonomous driving applications, it's of interest to confine the ego vehicle to travel within a particular region instead of allowing it to travel within the entire free space. This module achieves this by identifying a boundary of the region that the vehicle will be allowed to drive in, which may be retrieved from objects in the free space; curbs, lane lines, barriers, and walls. The boundary is converted to a series of evenly spaced virtual static nodes, with the spacing to be narrow enough such that the ego vehicle cannot cross between the virtual nodes.


\subsection{Triangulation Mesh Generator}\label{sec:mesh}
We employed \ac{dt} to generate a \ac{tm} from point representation of the static and dynamic objects extracted from the environment and constructs the corresponding dual graph. The dual graph has a node corresponding to each triangle of the mesh and an edge for each edge in the mesh that separates a pair of triangles. Target Attraction Principle~\cite{yan2008path} is used to position the dual graph nodes since the cost of reaching the triangles during graph search is computed based on the position of the dual graph node. \ac{cdt} or \ac{ddt} can also be used as alternatives to construct the \ac{tm}.


\subsection{Motion Information Transmission}
We achieve this by projecting the velocity $\boldsymbol{v}_i$ of each node $n_i$ onto the positional vectors $\boldsymbol{p}_{i,j}$ to their adjacent nodes $n_j$, where $\boldsymbol{p}_{i,j}$ is the positional vector from $n_i$ to $n_j$. We denote the projected velocity vector as $v'_i$. To preserve the heading information of $n_i$, $\boldsymbol{v}'_i$ will assume the direction of $\boldsymbol{v}_i$, thus $\boldsymbol{v}'_i =\lVert\boldsymbol{v}_i\rVert \cos{\theta}\hat{\boldsymbol{v}_i} = \cos{\theta}\boldsymbol{v}_i$. The effect of the proximity of $n_i$ to $n_j$ is accounted for by multiplying $\boldsymbol{v}'_i$ by $\frac{\alpha}{\rVert p \lVert + \alpha}$, $\alpha\geq 0$, so that $\lVert\boldsymbol{v}'_i\rVert$ decreases as $n_i$ is further away from $n_j$. The $\alpha$ is to balance the effect of velocity and proximity. Furthermore, the heading of node $n_i$ relative to $n_j$ is accounted for by $\boldsymbol{v}'_i$ by a term $|\frac{\pi}{2}-\theta|^\beta$, where $\theta \in [-\frac{\pi}{2},\frac{\pi}{2}]$, $\beta\geq0$, and $\theta$ is the angle between $\boldsymbol{v}_i$ and $\boldsymbol{p}_{i,j}$. If $\theta \notin [-\frac{\pi}{2},\frac{\pi}{2}]$, then we multiply $\boldsymbol{v}'_i$ by $0$ since $n_i$ is moving away from $n_j$. The $\beta$ is used to tune this term such that the effect of $\boldsymbol{v}_i$ diminishes as it becomes more orthogonal to~$\boldsymbol{p}_{i,j}$,

\vspace{-6pt}
\begin{equation}
\begin{aligned}
        \boldsymbol{v}'_i =
        \begin{cases}
                \frac{\alpha}{\rVert p \lVert + \alpha}|\frac{\pi}{2}-\theta|^\beta\cos{\theta}\boldsymbol{v}_i & \text{if} \ \theta \in [-\frac{\pi}{2},\frac{\pi}{2}] \\
                0 & \text{otherwise}
        \end{cases}
\end{aligned}.
\end{equation}

\begin{figure}[thpb]
	\centering
	\includegraphics[width=0.2\textwidth]{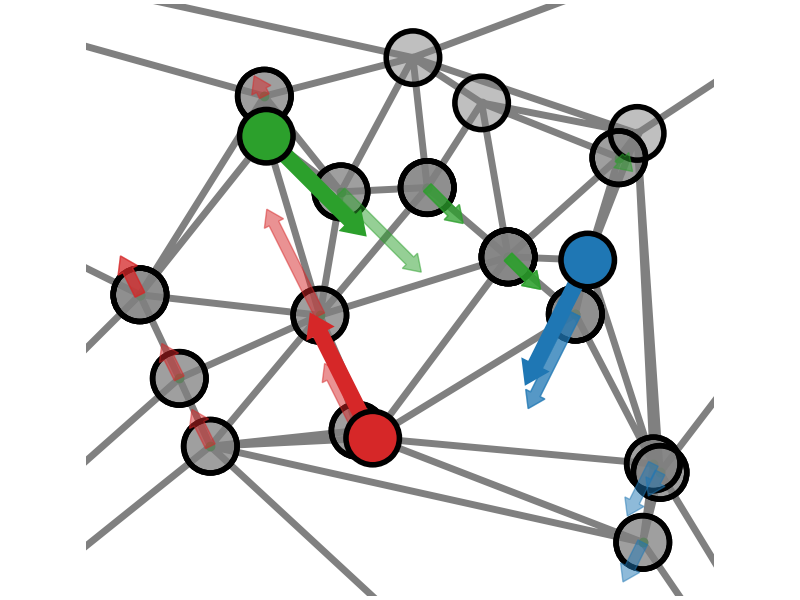}
	\vspace{-6pt}
	\caption{Motion information transmission example. Colored nodes are dynamic nodes, grey nodes are static nodes. All nodes with the same colored arrow received velocity information from the same node.}
	\label{fig:projection}
\end{figure}

To transmit the motion of the $n_i$ to $n_j$, we set the velocity vector $\boldsymbol{v}_j$ of $n_j$ to $\boldsymbol{v}'_i$ if $\lVert\boldsymbol{v}_j\rVert < \lVert\boldsymbol{v}'_i\rVert$ and that $\boldsymbol{v}_i\cdot\boldsymbol{p}_{i,j} > 0$, as expressed below,

\vspace{-6pt}
\begin{equation}
\begin{aligned}
        \boldsymbol{v}_j =
        \begin{cases}
                \boldsymbol{v}'_i & \text{if} \ \lVert\boldsymbol{v}_j\rVert < \lVert\boldsymbol{v}'_i\rVert \\
                \boldsymbol{v}_j            & \text{otherwise}
        \end{cases}
\end{aligned}.
\end{equation}
Figure~\ref{fig:projection} shows an example of applying this equation onto a set of nodes, where the velocity information of the dynamic nodes, red, green, and blue is projected onto the nodes around them.

\subsection{Graph Search}\label{sec:graphsearch}
This module finds an initial \ac{channel} that leads from start to goal within the \ac{tm} connectivity at $\tau_t$ by searching through the dual graph of \ac{tm}. We used Timed A*~\cite{cao2019dynamic}, which accounts for the path cost change due to object dynamics by computing $f(t_{eta}) = g(t_{eta}) + h(t_{eta})$, where \ac{teta} is the estimated time of arrival of the ego vehicle to a triangle. Additionally, it ensures that the distance between two nodes is wide enough to allow the ego vehicle to cross by checking \ac{teta} against the time interval in which the length of the corresponding \ac{dt} edge is above a threshold.

\subsection{Topological Event Prediction}\label{sec:compute}
This module determines $\tau_{t+1}$, the time of the first edge connectivity change of $\Delta_{t,m}$. Note that the triangle $\Delta_{t,e}$ contains the ego vehicle position, and we have limited our consideration to $e\leq m$. Therefore, we can efficiently find $\tau_{t+1}$ and $\Delta_{t,m}$ by iterating though $\Delta_{t,i} \in c_t$ to check if there is a change in edge connectivity before the time that the ego vehicle arrives at $\Delta_{t,i}$. If there is a change, $\Delta_{t,m} = \Delta_{t,i}$, otherwise $i = i+1$. We present this process in~Algorithm~\ref{algo:event}.

\subsubsection{In-circle Test}
The in-circle test~\cite{incircletestguibas1985primitives} defined in~\eqref{eqn:incircle test} is a standard test used to check if a 4\textsuperscript{th} point lies within the circumcircle of a triangle. This corresponds to the \texttt{incircleTest()} function in Algorithm~\ref{algo:event}.

The in-circle test is performed by calculating the determinant of a 4$\times$4 matrix, constructed from the position of the 4 nodes. Where $(x_i, y_i),~i\in \{1, 2, 3\}$ are the position of the triangle vertices. $(x_4, y_4)$ is the position of the 4$^{th}$ point.
The determinant of the 2\textsuperscript{nd} matrix corrects the signs of the in-circle test such that it is insensitive to the ordering of the first three points. If $\gamma = 0$, the 4\textsuperscript{th} point is co-circular with the circumcircle, where $\gamma < 0 $ and $\gamma > 0 $, respectively, indicate that the 4\textsuperscript{th} point is inside or outside the circumcircle.

\vspace{-6pt}
\begin{equation}
	\gamma =
	\det{\begin{bmatrix}
		1 & x_1 & y_1 & x_1^2 + y_1^2 \\
		1 & x_2 & y_2 & x_2^2 + y_2^2 \\
		1 & x_3 & y_3 & x_3^2 + y_3^2 \\
		1 & x_4 & y_4 & x_4^2 + y_4^2 
		\end{bmatrix}}
	\times
	\det{\begin{bmatrix}
		1 & x_1 & y_1 \\
		1 & x_2 & y_2 \\
		1 & x_3 & y_3
		\end{bmatrix}}
	\label{eqn:incircle test}
\end{equation}

\subsubsection{Compute Event Time} 
Algorithm~\ref{algo:event} computes $\tau_{t+1}$ by performing the in-circle test on each $\Delta_{t,i} \in c_t$ using $(x_i(t), y_i(t)), i\in \{1,\dots,4\}$, where $t \in [\tau_t, t_{\text{eta}}]$. $(x_i(t), y_i(t))$ is the expected position of the vertices of $\Delta_{t,i}$ and the point tested by the in-circle test at time $t$. Time $t$ is a time point sampled at even intervals from $\tau_t$ to $t_{\text{eta}}$. The $t_{\text{eta}}$ is the ego vehicle's estimated time of arrival at $\Delta_{t,i}$ because we only consider the topological event that occurred before the ego vehicle arrives at $\Delta_{t,i}$.

\begin{algorithm}\label{algo:event}
    \footnotesize
	\DontPrintSemicolon
	\SetKwFunction{FMain}{main}
	\SetKwFunction{FVert}{vertices}
	\SetKwFunction{FEta}{getETA}
	\SetKwFunction{Fadj}{getNeighbors}
	\SetKwFunction{Fmodel}{dynamicModel}
	\SetKwFunction{Ftest}{incircleTest}
	\SetKwProg{Def}{def}{:}{}
				
	\KwIn{$c_t$, sample resolution}
	\KwOut{$\tau_{t+1}, \Delta_{t,i}$}
			
	\Def{\FVert{$\Delta_{t,i}$}}{	
		\KwRet{vertices of $\Delta_{t,i}$}
	}
			
	\Def{\FEta{$\Delta_{t,i}$}}{	
		\KwRet{$t_{eta}$}
	}
			
	\Def{\Fadj{$\Delta_{t,i}$}}{	
		\KwRet{All vertices of the adjacent triangles that are not also vertices of $\Delta_{t,i}$}
	}
	\Def{\Fmodel{p1, p2, p3, p4, t}}{	
		\KwRet{position of input points at time t}
	}
				
	\Def{\Ftest{a, b, c, p}}{	
		\KwRet{$\gamma < 0$}
	}
			
	\Def{\FMain{$c_t$, sample resolution}}{
		\For{$\Delta_{t,i}$ in $c_i$}{
			a, b, c = \FVert{$\Delta_{t,i}$}\;
			$t_{eta}$ = \FEta{$\Delta_{t,i}$}\;
												 
			\For{p in \Fadj{$\Delta_{t,i}$}}{
				\For{$\tau$ in range (0, $t_{eta}$, sample resolution)}{
					
					a, b, c, p = \Fmodel{a, b, c, p $\tau$}		
						
					\If{\Ftest{a, b, c, p}}{
						\KwRet{$\tau, \Delta_{t,i}$}
					}
				}
			}
									
		}
		\KwRet None\;
	}
	\caption{Compute Topological Event Time}
\end{algorithm}

Algorithm~\ref{algo:event} terminates if the in-circle test finds a topological event, in which $\tau_{t+1}$ is the returned time and $\Delta_{t,m}$ is the returned triangle. Algorithm 1 also terminates if no event is found for any triangle within the channel, which indicates that \ac{channel} is the final segment of \ac{finalchannel}.

\subsubsection{Dynamic Object Motion Model}
This module corresponds to the \texttt{dynamicModel()} function in Algorithm~\ref{algo:event}. This returns $(x_i(t_j), y_i(t_j))$, which is a function that estimates the position of nodes at time $t_j$, $t_j \in [\tau, t_{\text{eta}}]$, $\mathbf{p^i(t)} =\mathbf{p^i_0} + f(v, t)$,
where $f(v, t)$ is a linear or nonlinear function. $v$ is the current velocity of the object, and $t$ is time. In our implementation, we used a linear model, \textit{i.e.}~$f(v, t) = \mathbf{v^it}$, to estimate the position of the nodes.

\subsubsection{Estimate Ego Arrival Time}
This module corresponds to the \texttt{getETA()} function in Algorithm~\ref{algo:event}. It estimates the time of arrival of the ego vehicle $t_{eta}$ to $\Delta_{t,i}$ following the linear motion model. The Target Attraction Principle proposed by~\cite{yan2008path} is used to position the dual graph node that represents each triangle. $t_{eta}$ is estimated by taking the ego vehicle's speed over the travel distance from the starting position to the dual graph node that corresponds to $\Delta_{t,i}$.

\subsection{Generate Channel Segment}
Module~\ref{sec:IDanchor} identifies $c'_t$ by finding triangle $\Delta_{t,k} \in c_t$, which is the last triangle of $c'_t$. Additionally, since path planners typically requires a point to navigate to, module~\ref{sec:subgoal} identifies a subgoal point within $\Delta_{t,k}$ for path planning.

\subsubsection{Channel Segment's Last Triangle Identifier}\label{sec:IDanchor}
$\Delta_{t,k}$ is found with the following:
\begin{itemize}
        \item if $e<m$, then $\Delta_{t,k}$ = $\Delta_{t,e}$ , where $m$ is the index of $\Delta_{t,m} \in c_t$;
        \item if $e \geq m$, then $\Delta_{t,k}$ = $\Delta_{t,m-1}$.
\end{itemize}

To spatially and temporally connect $c'_t$ to $c'_{t+1}$, we consider the last triangle of $c'_t$ to also be the first triangle of $c'_{t+1}$, $\Delta_{t,k}=\Delta_{t+1,0}$. Therefore, $\Delta_{t,k}$ serves as the triangle that transitions from $c'_t$ to $c'_{t+1}$ at time $\tau_{t+1}$.

\subsubsection{Sub-Goal Point Generation}\label{sec:subgoal}
Since local planner typically requires a point to navigate to, this module chooses a collision-free point inside the $\Delta_{t,k}$, that is closest to the estimated ego vehicle position at \ac{tevent}. 

\subsection{Terminal Condition}
Algorithm~\ref{algo:event} will return None when the input \ac{channel} leads to the goal point without any of the triangles in \ac{channel} seeing a topological event. This means that $C = \{c'_0, \dots, c_t\}$, where each channel $c \in C$ can be used to plan a path segment.

Otherwise, our system finds $c'_{t+1}$ by repeating from module~\ref{sec:mesh} using the expected position of the starting node and the dynamic objects at $\tau_{t+1}$ found by module~\ref{sec:compute}. Our system as shown in Figure~\ref{fig:process diagram} may also choose to terminate early to limit the planning horizon when the first $n$ channel segments $c'$ is found, or when the predicted $\tau_{t+1} > \tau_{\text{threshold}}$.

\begin{figure}[t]
	\centering
	\includegraphics[width=0.35\textwidth]{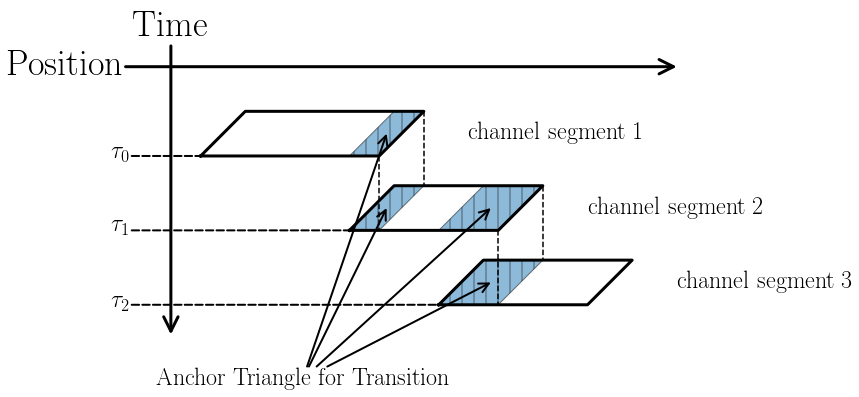}
	\vspace{-6pt}
	\caption{Conceptual diagram of the overall generated channel \ac{finalchannel}. The anchor triangles mark the overlapping triangles in each transition.}
	\label{fig:channel concept}
\end{figure}

%
Figure~\ref{fig:channel concept} demonstrates a conceptual understanding of \ac{finalchannel}.
Each $c'_t$ is valid for from time $\tau_t$ until the next time of topological event $\tau_{t+1}$. To transition from one channel to the next, the last triangle $\Delta_{t,k}$ of $c'_t$ will overlap with the first triangle of the next channel segment $c'_{t+1}$.

\section{EXPERIMENTS}\label{sec:experiment}
We carried out two experiments to verify the effectiveness of our proposed method against a synthetic set of scenarios and a set of real world scenarios from the nuScenes dataset~\cite{nuscenes2019}. We compare the channels generated by our proposed method against the channels generated by existing triangulation-based planning methods that uses A* and Timed A* to search for the channels. Because there are no metrics to directly evaluate channels, the channels are evaluated based on the performance of the path planned using those channels. The Funnel Algorithm~\cite{funnelalgorithm} is used to produce the path using the channel generated by each method. We add an offset to the vertices of the funnel as padding for collision avoidance. Note that the simplicity of the funnel algorithm cannot guarantee the path to be collision free. The channels are compared in terms of rate of task completion, the time taken to complete each task, rate of successfully planning a path, and rate of collision. All experiments are implemented using Python and carried out on a computer with a 3.6GHz CPU and 32GB of memory. 

\subsection{Scenarios}
\subsubsection{Synthetic}
This experiment is a set of 200 synthetic scenarios to test the long-term planning capability of our proposed method by introducing frequent changes in the triangulation mesh connectivity. These scenarios have between 10 to 20 pedestrians with randomized position and speed, ranging from 0.25~m/s to 1~m/s crossing perpendicularly to a 30~m straight road. The ego vehicle is tasked to navigate to the end of the road at 2~m/s within a simulation time limit of 25~s. 

\subsubsection{nuScenes}
This experiment is a set of 150 real-world scenarios from the nuScenes dataset~\cite{nuscenes2019} to test the proposed method's performance in realistic scenarios. These scenarios are collected from typical driving conditions such as as highway, intersections, and parking lots.
Each scenario is 20~s long, annotated at 2~Hz. We linearly interpolated the position of objects in the scenario at 10~Hz for higher temporal resolution. All dynamic objects' motion after the annotated 20~s is assumed to be linear. The ego vehicle is tasked to navigate, starting from the data collection vehicle's pose at $t=0$ to the pose at $t=20$ at the average velocity of the data collection vehicle within 30~s in simulation time. The experiment environment is a rectangular region that encloses the start and goal point.
We confine the ego vehicle's motion to the road by placing virtual static nodes on the boundary of the drivable area map layer of the nuScenes dataset. On average, each scenario contains 389 nodes, where 55 of which are dynamic or static objects identified from the environment.

%
%
%
\subsection{Result}
\begin{figure}[h]
	\centering
	\includegraphics[width=0.49\textwidth]{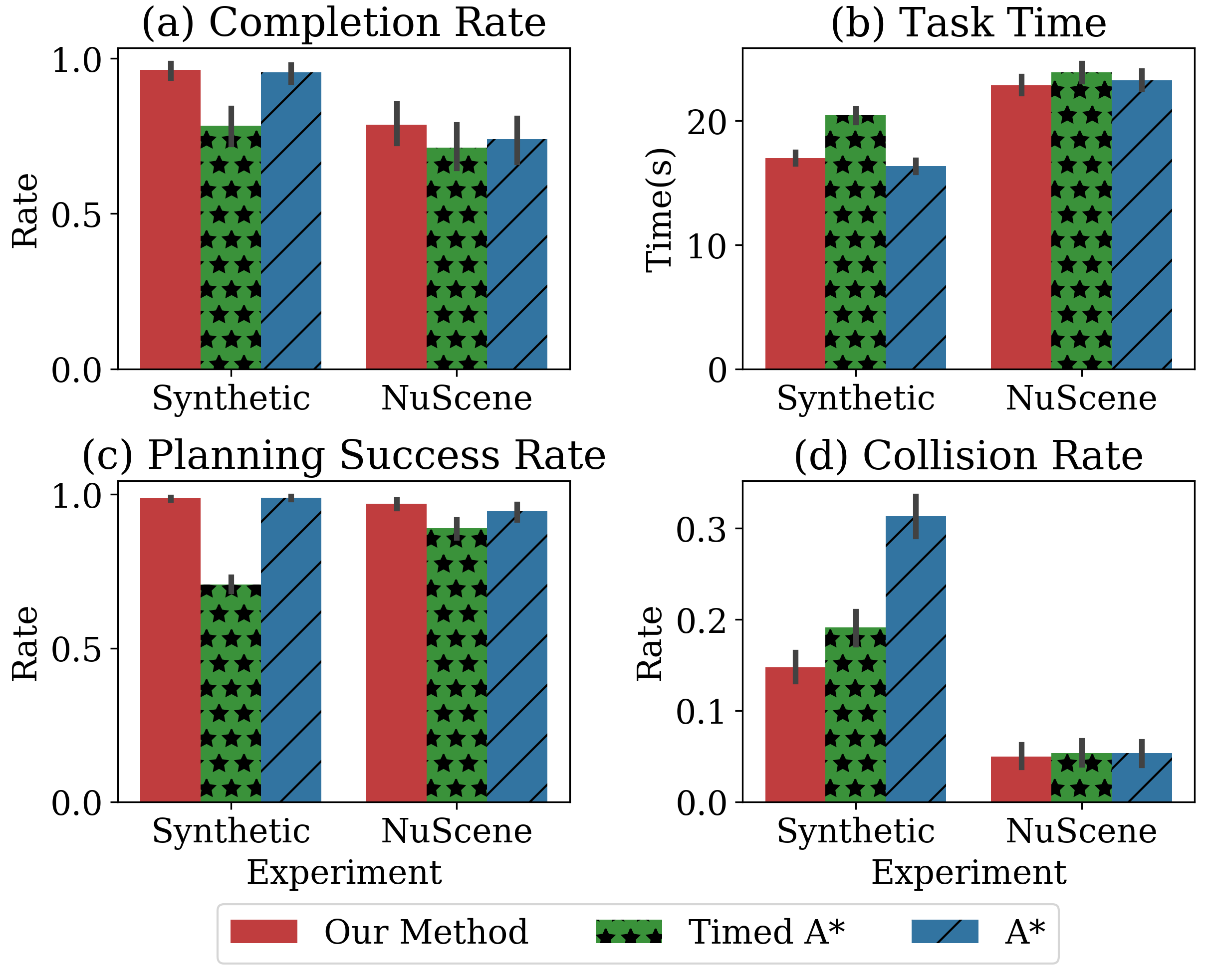}
	\vspace{-6pt}
	\caption{Experimental results in comparison to Timed A*~\cite{cao2019dynamic} and A* against a Synthetic dataset and a real-world dataset from nuScenes.}
	\label{fig:straight}
\end{figure}

This section presents the comparison results obtained based on the Synthetic and nuScenes scenarios. A* is expected to result in a higher completion rate, higher success rate, and lower task time as it ignores the dynamic objects. Because A* ignores dynamic objects, it is also expected to result in a higher collision rate. As such, a desirable method, ideally, should have comparable completion rate, success rate, and task time to those of A*, while resulting in a lower collision rate. Figure~\ref{fig:straight} summarizes the comparison results between A*, Timed A*~\cite{cao2019dynamic}, and our method against a Synthetic dataset and a real-world dataset from nuScenes.
 
\subsubsection{Synthetic}
Our method completed 96.3\% of tasks at an average time of 17~s, Timed A* completed 78.6\% of the tasks at an average time of 21~s, and A* completed 95.5\% of the task at 16.3~s. Our method produced spatial constraints that allowed the local planner to find a valid plan in 98.7\% of the planning cycles, whereas Timed A* was 70.7\% and A* at 98.9\%.

The Timed A* based method has a lower success rate of finding a path at each planning cycle. This is because Timed A* looks for a channel based on the future state of the triangulation mesh, while it provides the path planner with a channel at the present state. The channel width at the present state may not be large enough to allow the ego vehicle to pass, thus the path planner fails. The path planner failure cases for both our method and A* based method are situations where the ego vehicle needed to stop and wait for dynamic objects in front of it to pass before it can proceed.


\begin{figure}[bthp]
	\centering
	\subfigure[Time = 0.00 s]{%
	    \setlength{\fboxrule}{0.7pt}%
		\fbox{\includegraphics[width=0.22\textwidth]{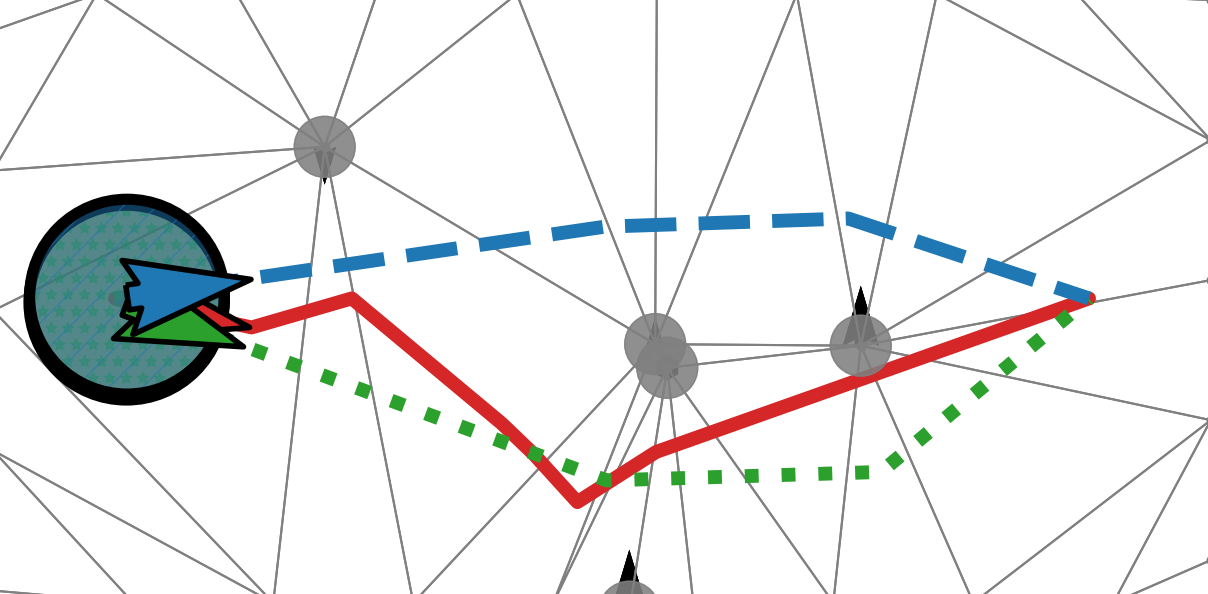}}}
		\vspace{-4pt}
	\subfigure[Time = 3.60 s]{%
	    \setlength{\fboxrule}{0.7pt}%
		\fbox{\includegraphics[width=0.22\textwidth]{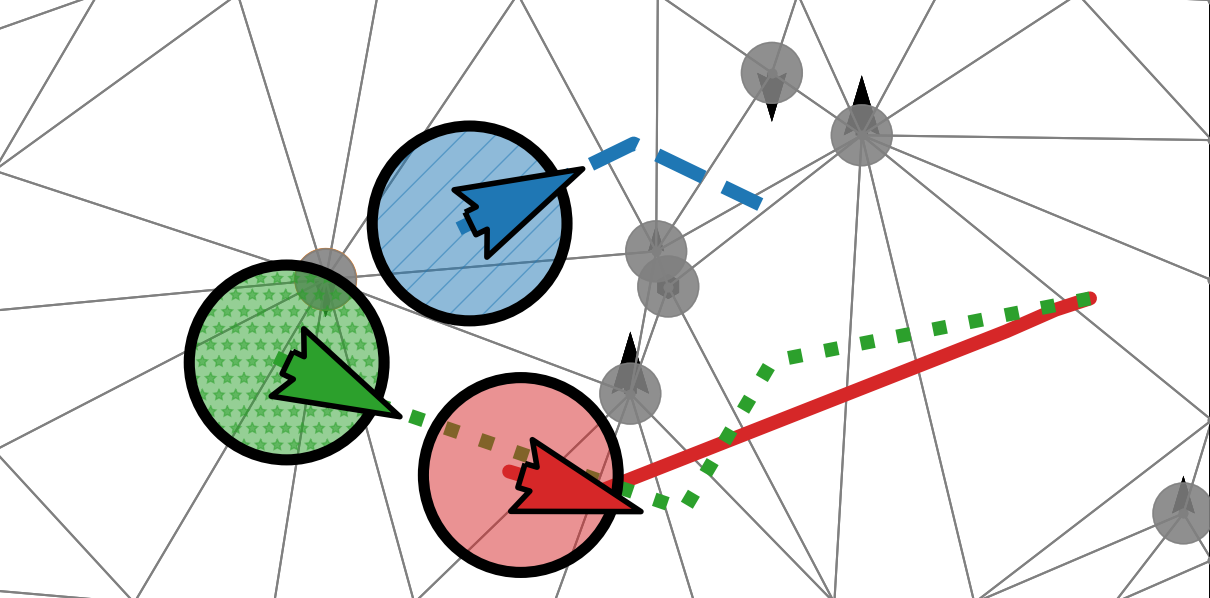}}} \\
		\vspace{-4pt}
	\subfigure[Time = 4.00 s]{%
	    \setlength{\fboxrule}{0.7pt}%
		\fbox{\includegraphics[width=0.22\textwidth]{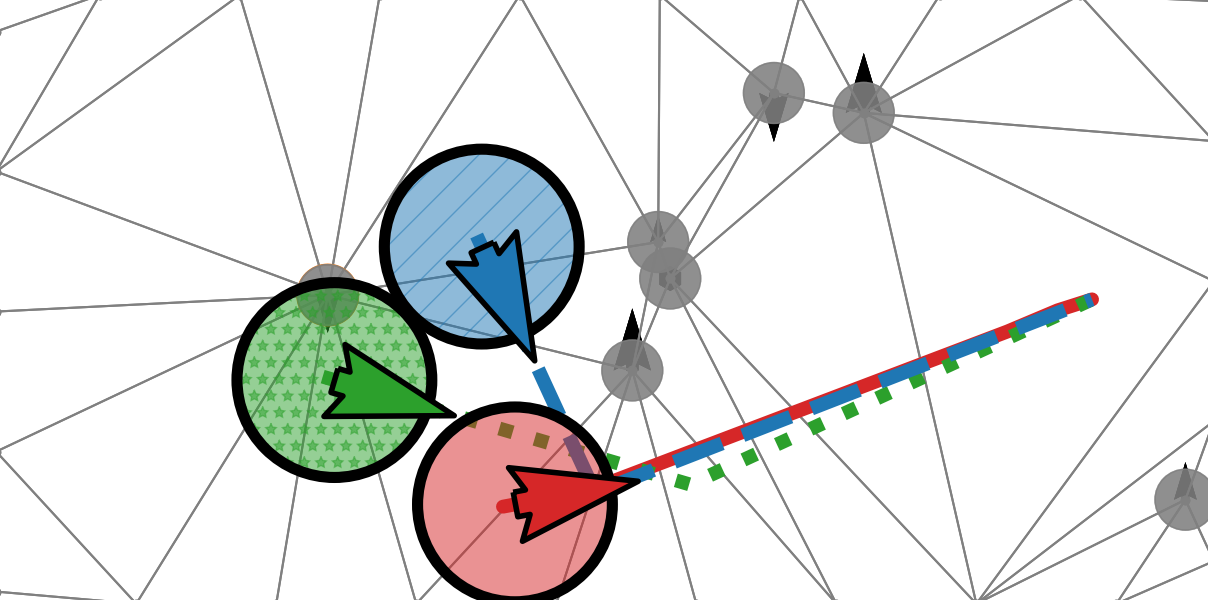}}}
	\subfigure[Time = 9.00 s]{%
	    \setlength{\fboxrule}{0.7pt}%
		\fbox{\includegraphics[width=0.22\textwidth]{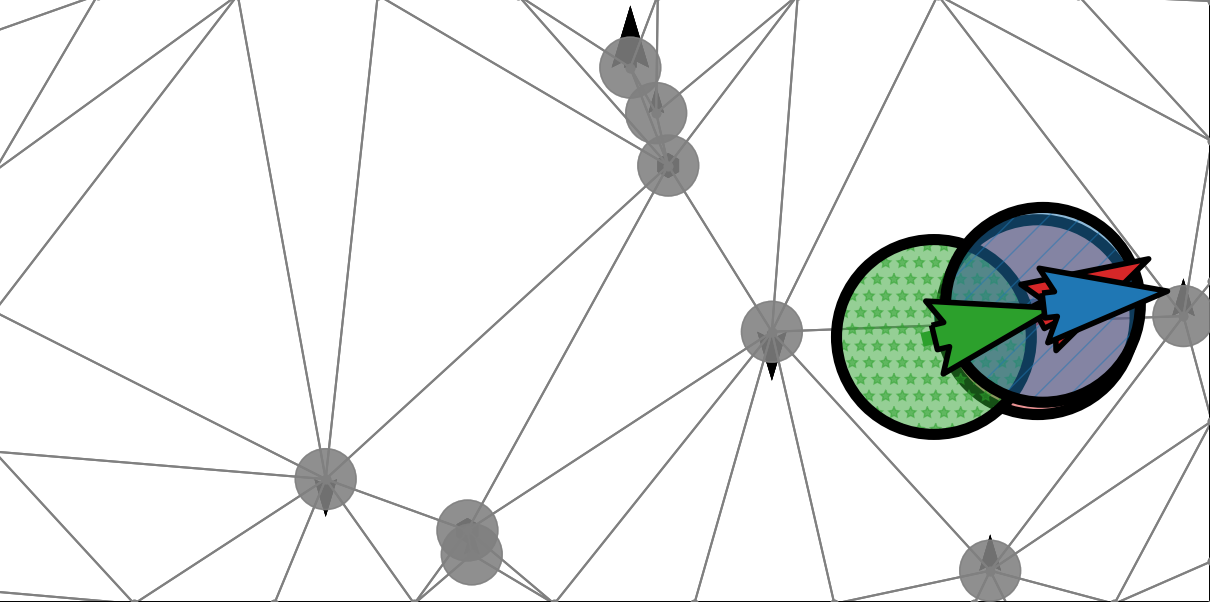}}}
	\caption{Straight Scenario Example. Our method (red solid), Timed A* (green dotted), A* (blue dashed). Grey circles (pedestrians).}
	\label{fig:straight_2}
\end{figure}


\begin{figure}[thbp]
	\centering
	\vspace{-4pt}
	\subfigure[Time = 0.0 s]{%
	    \setlength{\fboxrule}{0.7pt}%
		\fbox{\includegraphics[trim=2.5cm 2.5cm 2.5cm 2.6cm, clip, width=0.4\textwidth]{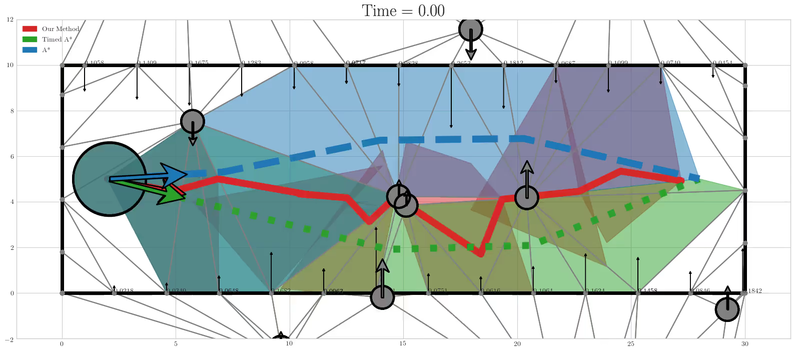}}
		\label{fig:straight_1_1}} \\
	\vspace{-4pt}
	\subfigure[Time = 4.30 s]{%
	    \setlength{\fboxrule}{0.7pt}%
		\fbox{\includegraphics[trim=2.5cm 2.5cm 2.5cm 2.6cm, clip, width=0.4\textwidth]{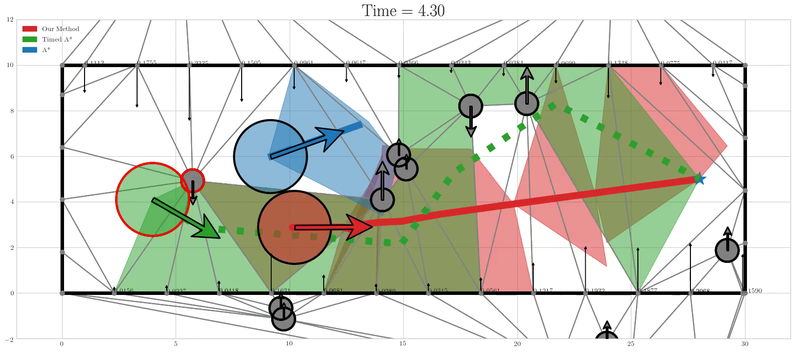}}
		\label{fig:straight_1_2}} \\
	\vspace{-4pt}
	\subfigure[Time = 7.30 s]{%
	    \setlength{\fboxrule}{0.7pt}%
		\fbox{\includegraphics[trim=2.5cm 2.5cm 2.5cm 2.6cm, clip, width=0.4\textwidth]{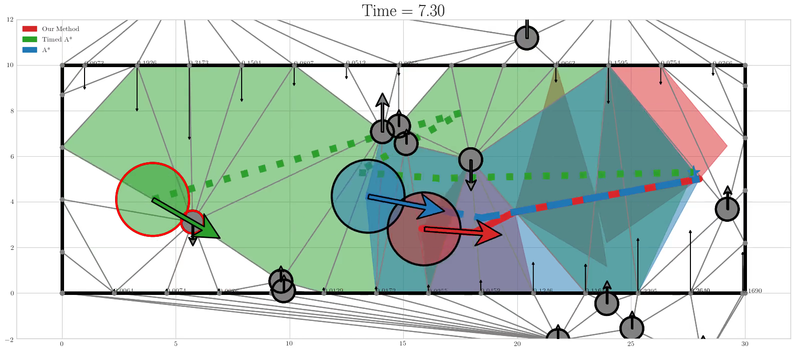}}
		\label{fig:straight_1_3}} \\
	\vspace{-4pt}
		\subfigure[Time = 10.30 s]{%
	    \setlength{\fboxrule}{0.7pt}%
		\fbox{\includegraphics[trim=2.5cm 2.5cm 2.5cm 2.6cm, clip, width=0.4\textwidth]{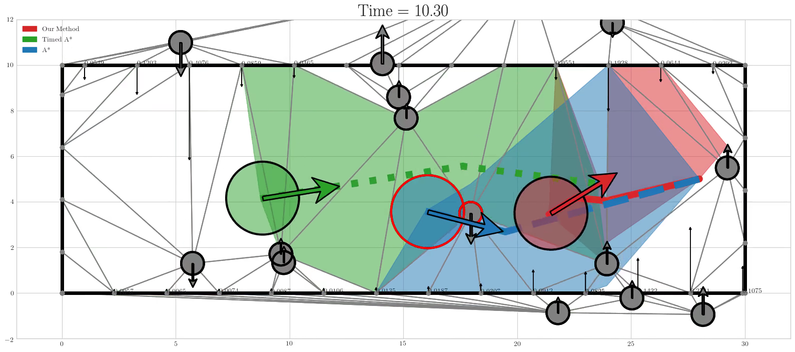}}
		\label{fig:straight_1_4}} \\
	\vspace{-4pt}
	\subfigure[Time = 10.30 s]{%
	    \setlength{\fboxrule}{0.7pt}%
		\fbox{\includegraphics[trim=2.5cm 2.5cm 2.5cm 2.6cm, clip, width=0.4\textwidth]{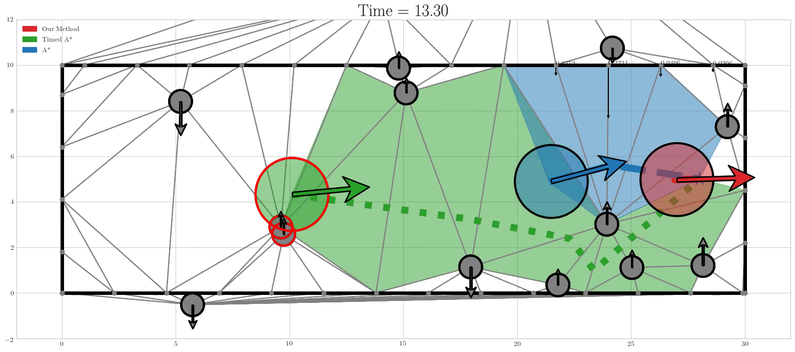}}
		\label{fig:straight_1_5}} \\
	\caption{A sample scenario, in which both A* and Timed A* fail. Our method (red solid), Timed A* (green dotted), A* (blue dashed). Grey circles (pedestrians). The colliding objects are marked by red circumference.}
	\label{fig:straight_1}
\end{figure}

Since A* ignores the motion of dynamic objects when it generates the spatial constraint, it may result in unfavorable paths that are not suitable for autonomous driving. The effect of this is evident in its high collision rate of 31.3\%, whereas our method is at 14.8\% and Timed A* at 19.1\%. Figure~\ref{fig:straight_2} shows an example of the unfavorable situation, where the A* method shown by the blue dashed line resulted in a path that ignored the motion of the dynamic object until $t=3.6$. It was then forced to make a sudden course change at $t=4.0$ because the previous path was no longer feasible. The path generated by our method shown in the red solid line and Timed A* shown in the green dotted line both moved in another direction to account for the pedestrian motion and have anticipated that the path will not be feasible at a later time. Figure~\ref{fig:straight_1} also shows a similar scenario, in which both Timed A* and A* collide with the crossing objects due to invalid triangulations and lack of collision checks for dynamic objects, respectively.
Note that some of the collisions are contributed by the simplicity of the path planner not providing sufficient spacing when passing an object. A more sophisticated local planner would result in fewer or no collisions.

\subsubsection{nuScenes}
\begin{figure}[bthp]
	\centering
	\subfigure[Time = 2.80 s]{%
	    \setlength{\fboxrule}{0.7pt}%
		\fbox{\includegraphics[trim=7cm 12.5cm 3cm 2.6cm, clip, width=0.2\textwidth]{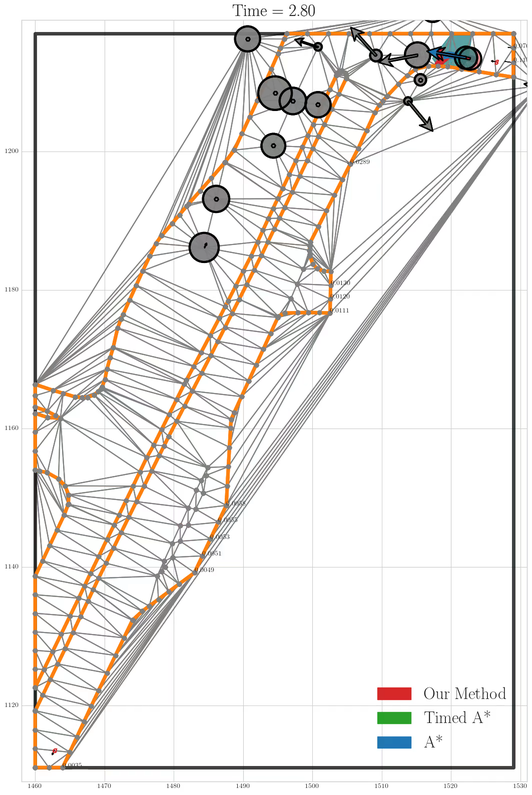}}
		\label{fig:nuscene_sparse_1}}
	\subfigure[Time = 7.30 s]{%
	    \setlength{\fboxrule}{0.7pt}%
		\fbox{\includegraphics[trim=7cm 12.5cm 3cm 2.6cm, clip, width=0.2\textwidth]{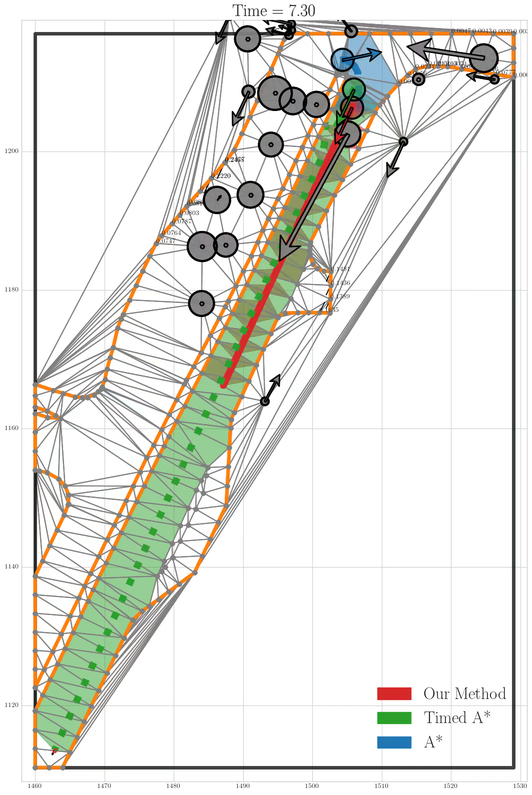}}
		\label{fig:nuscene_sparse_2}} \\
	\vspace{-4pt}
	\subfigure[Time = 13.30 s]{%
	    \setlength{\fboxrule}{0.7pt}%
		\fbox{\includegraphics[trim=7cm 12.5cm 3cm 2.6cm, clip, width=0.2\textwidth]{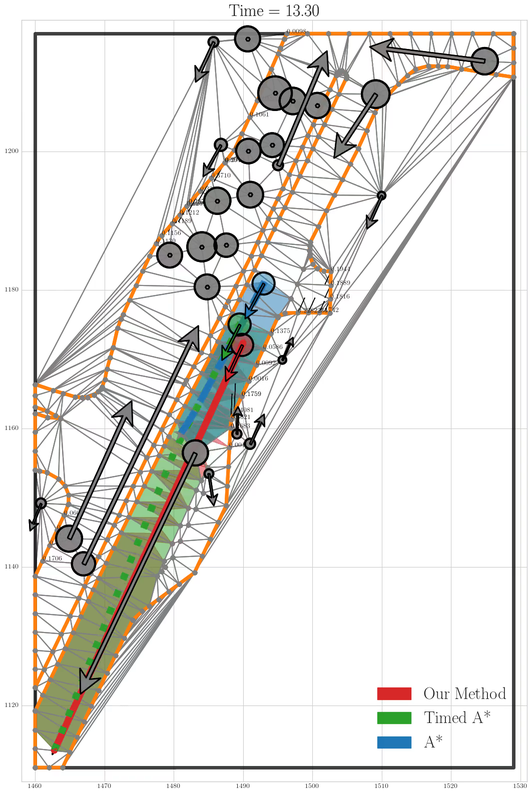}}
		\label{fig:nuscene_sparse_3}}
	\subfigure[Time = 17.80 s]{%
	    \setlength{\fboxrule}{0.7pt}%
		\fbox{\includegraphics[trim=7cm 12.5cm 3cm 2.6cm, clip, width=0.2\textwidth]{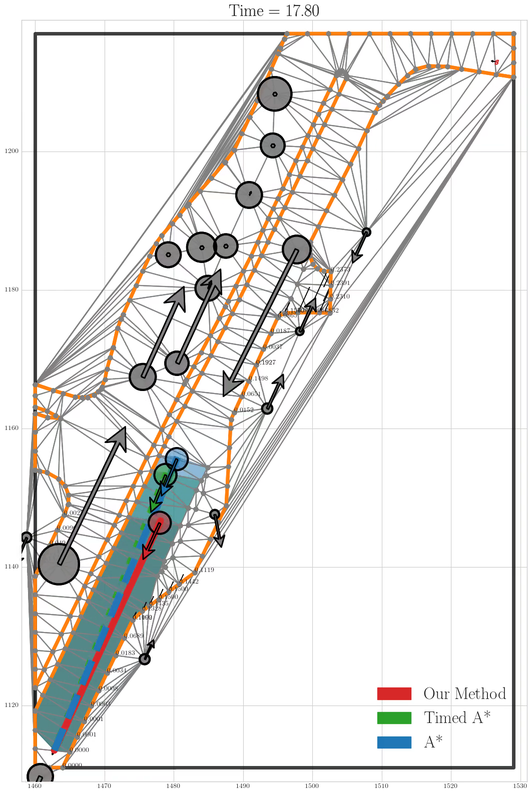}}
		\label{fig:nuscene_sparse_4}}
	\caption{A Sample nuScenes Lane Following Scenario. Our method (red solid), Timed A* (green dotted), A* (blue dashed). Grey circles (dynamic or static objects).}
	\label{fig:nuscene_sparse}
\end{figure}

Our method completed the task 78.7\%, outperforming Timed A*, which was at 71.3\% and A* at 74.1\%. The average time of completion is similar for all three methods; 22.9~s for our method, 23.9~s for Timed A*, and 23.25~s for A*. Using the channel produced by our method, the local planner was able to find a path 97\% of the time, where it was 89.1\% for Timed A* and 94.5\% for A*. Because most scenarios are similar to a lane following task, the collision rate is similar for all three methods; 5\% for our method, and 5.3\% for both Timed A* and A*. This is expected as pedestrians or other objects mostly move along the road in these scenarios, which is an easier motion planning task compared to pedestrian crossing or lane-changing scenarios. This is also shown in Figure~\ref{fig:nuscene_sparse}, in which the objects do not cross the lane for a greater portion of the scenario.

\begin{figure}[bthp]
	\centering
	\subfigure[Time = 5.30 s]{%
	    \setlength{\fboxrule}{0.7pt}%
		\fbox{\includegraphics[trim=0.4cm 11.5cm 0.4cm 0.4cm, clip, width=0.49\textwidth]{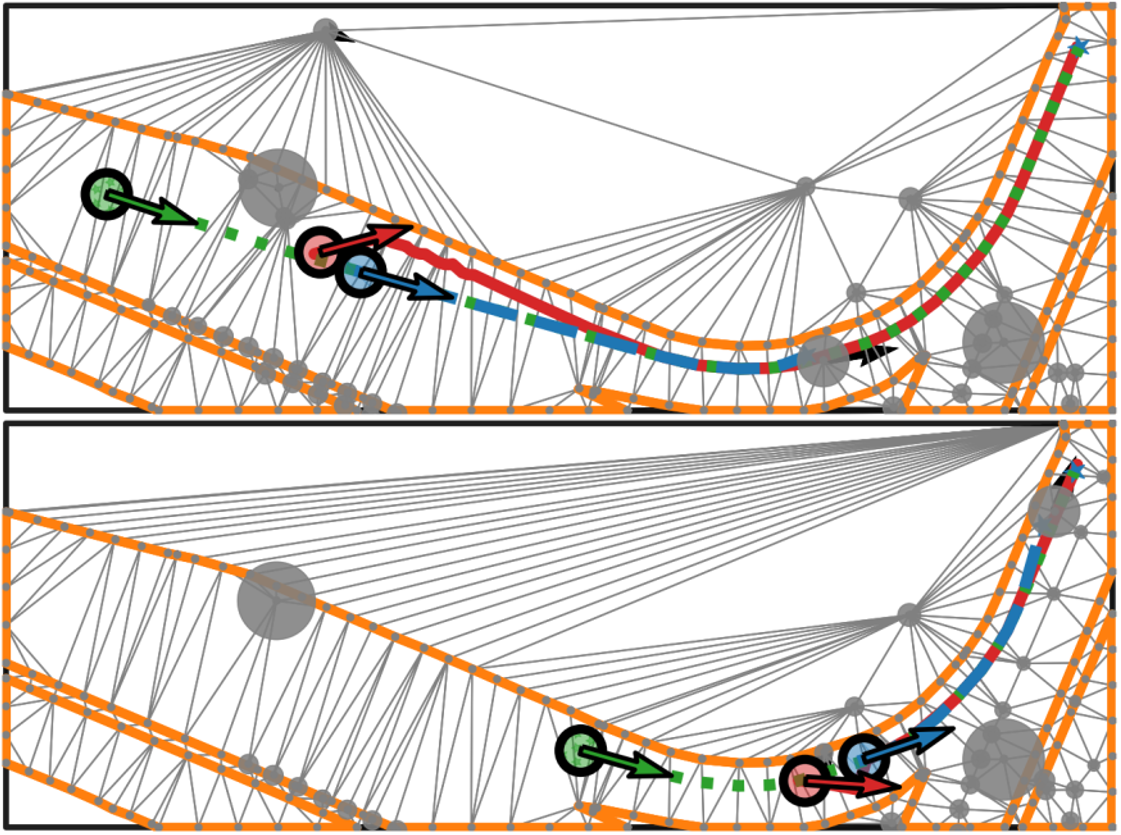}}
		\label{fig:experiment_fig_top}} \\
	\vspace{-4pt}
	\subfigure[Time = 13.70 s]{%
		\setlength{\fboxrule}{0.7pt}%
		\fbox{\includegraphics[trim=0.4cm 0.4cm 0.4cm 11.5cm, clip, width=0.49\textwidth]{assets/experiment_fig_top.png}}
		\label{fig:experiment_fig_bott}} 
	\caption{nuScenes Scenario Example. Our method (red solid), Timed A* (green dotted), A* (blue dashed). Grey circles (dynamic or static objects).}
	\label{fig:nuScenes}
\end{figure}


Figure~\ref{fig:nuScenes} shows an example of the nuScenes scenario where Timed A* identified a feasible channel, however, the channel was based on the current triangulation mesh connectivity. This channel did not have sufficient space to allow the local planner to find a path, thus the local planner was unable to find a path until $t=5.3$. At $t=13.4$, a pedestrian was about to cross the road, our proposed method accounted for this when generating a channel and tried to maneuver around the pedestrian, as shown in the slight turn in its path. In contrast, A* would've only stopped when the spacing to cross becomes too narrow.
\section{CONCLUSIONS}\label{conclusion}

In this paper, we presented a novel method to generate spatial constraints for motion planning in dynamic environments. 
This method can enhance other planning modules. For example, it can act as a preprocessor to generate spatial constraints as inputs or to prune candidate trajectories generated by the motion planner. It can perform mission planning to find a global route or to use it as a safeguard against semantic map failures that may be caused by dynamic changes in the map (e.g.\ road obstruction and construction zones). Furthermore, the proposed system's modularity allows for adding, removing, and replacing the various modules as desired.
We evaluated our method in a set of real-world scenarios simulated based on an autonomous driving dataset. The result shows that using our method can obtain a more stable, long-term plan that yields a higher task completion rate, faster travel time, higher planning success rate, and fewer collisions compared to other existing methods.

\bibliographystyle{IEEEtran}
\bibliography{IEEEexample}

\begin{thebibliography}{10}
\providecommand{\url}[1]{#1}
\csname url@samestyle\endcsname
\providecommand{\newblock}{\relax}
\providecommand{\bibinfo}[2]{#2}
\providecommand{\BIBentrySTDinterwordspacing}{\spaceskip=0pt\relax}
\providecommand{\BIBentryALTinterwordstretchfactor}{4}
\providecommand{\BIBentryALTinterwordspacing}{\spaceskip=\fontdimen2\font plus
\BIBentryALTinterwordstretchfactor\fontdimen3\font minus
  \fontdimen4\font\relax}
\providecommand{\BIBforeignlanguage}[2]{{%
\expandafter\ifx\csname l@#1\endcsname\relax
\typeout{** WARNING: IEEEtran.bst: No hyphenation pattern has been}%
\typeout{** loaded for the language `#1'. Using the pattern for}%
\typeout{** the default language instead.}%
\else
\language=\csname l@#1\endcsname
\fi
#2}}
\providecommand{\BIBdecl}{\relax}
\BIBdecl

\bibitem{kallmann2005path}
M.~Kallmann, ``Path planning in triangulations,'' in \emph{Proceedings of the
  IJCAI workshop on reasoning, representation, and learning in computer games},
  2005, pp. 49--54.

\bibitem{chen2010enhanced}
J.~Chen, C.~Luo, M.~Krishnan, M.~Paulik, and Y.~Tang, ``An enhanced dynamic
  delaunay triangulation-based path planning algorithm for autonomous mobile
  robot navigation,'' in \emph{Intelligent Robots and Computer Vision XXVII:
  Algorithms and Techniques}, vol. 7539.\hskip 1em plus 0.5em minus 0.4em\relax
  International Society for Optics and Photonics, 2010, p. 75390P.

\bibitem{yan2008path}
H.~Yan, H.~Wang, Y.~Chen, and G.~Dai, ``Path planning based on constrained
  delaunay triangulation,'' in \emph{2008 7th World Congress on Intelligent
  Control and Automation}.\hskip 1em plus 0.5em minus 0.4em\relax IEEE, 2008,
  pp. 5168--5173.

\bibitem{kallmann2014navigation}
M.~Kallmann and M.~Kapadia, ``Navigation meshes and real-time dynamic planning
  for virtual worlds,'' in \emph{ACM SIGGRAPH 2014 Courses}, 2014, pp. 1--81.

\bibitem{perkins2013field}
S.~Perkins, P.~Marais, J.~Gain, and M.~Berman, ``{Field D*} path-finding on
  weighted triangulated and tetrahedral meshes,'' \emph{Autonomous agents and
  multi-agent systems}, vol.~26, no.~3, pp. 354--388, 2013.

\bibitem{demyen2006efficient}
D.~Demyen and M.~Buro, ``Efficient triangulation-based pathfinding,'' in
  \emph{Aaai}, vol.~6, 2006, pp. 942--947.

\bibitem{broz2014dynamic}
P.~Broz, M.~Zemek, I.~Kolingerov{\'a}, and J.~Szkandera, ``Dynamic path
  planning with regular triangulations,'' \emph{Machine Graphics \& Vision},
  vol.~24, no. 3/4, pp. 119--142, 2014.

\bibitem{devillers1992fully}
O.~Devillers, S.~Meiser, and M.~Teillaud, ``Fully dynamic delaunay
  triangulation in logarithmic expected time per operation,''
  \emph{Computational Geometry}, vol.~2, no.~2, pp. 55--80, 1992.

\bibitem{mostafavi2003delete}
M.~A. Mostafavi, C.~Gold, and M.~Dakowicz, ``Delete and insert operations in
  {Voronoi/Delaunay} methods and applications,'' \emph{Computers \&
  Geosciences}, vol.~29, no.~4, pp. 523--530, 2003.

\bibitem{chew1989constrained}
L.~P. Chew, ``Constrained {Delaunay} triangulations,'' \emph{Algorithmica},
  vol.~4, no. 1-4, pp. 97--108, 1989.

\bibitem{edelsbrunner1996incremental}
H.~Edelsbrunner and N.~R. Shah, ``Incremental topological flipping works for
  regular triangulations,'' \emph{Algorithmica}, vol.~15, no.~3, pp. 223--241,
  1996.

\bibitem{likhachev2005anytime}
M.~Likhachev, D.~I. Ferguson, G.~J. Gordon, A.~Stentz, and S.~Thrun, ``{Anytime
  Dynamic A*}: An anytime, replanning algorithm.'' in \emph{ICAPS}, vol.~5,
  2005, pp. 262--271.

\bibitem{koenig2002d}
S.~Koenig and M.~Likhachev, ``{D* Lite},'' \emph{Aaai/iaai}, vol.~15, 2002.

\bibitem{ferguson2005field}
D.~Ferguson and A.~Stentz, ``The {Field D*} algorithm for improved path
  planning and replanning in uniform and non-uniform cost environments,''
  \emph{Robotics Institute, Carnegie Mellon University, Pittsburgh, PA, Tech.
  Rep. CMU-RI-TR-05-19}, 2005.

\bibitem{likhachev2004ara}
M.~Likhachev, G.~J. Gordon, and S.~Thrun, ``{ARA*}: {Anytime A*} with provable
  bounds on sub-optimality,'' in \emph{Advances in neural information
  processing systems}, 2004, pp. 767--774.

\bibitem{cao2019dynamic}
C.~Cao, P.~Trautman, and S.~Iba, ``Dynamic channel: A planning framework for
  crowd navigation,'' in \emph{2019 International Conference on Robotics and
  Automation (ICRA)}.\hskip 1em plus 0.5em minus 0.4em\relax IEEE, 2019, pp.
  5551--5557.

\bibitem{incircletestguibas1985primitives}
L.~Guibas and J.~Stolfi, ``Primitives for the manipulation of general
  subdivisions and the computation of {Voronoi},'' \emph{ACM transactions on
  graphics (TOG)}, vol.~4, no.~2, pp. 74--123, 1985.

\bibitem{nuscenes2019}
H.~Caesar, V.~Bankiti, A.~H. Lang, S.~Vora, V.~E. Liong, Q.~Xu, A.~Krishnan,
  Y.~Pan, G.~Baldan, and O.~Beijbom, ``{nuScenes}: A multimodal dataset for
  autonomous driving,'' \emph{arXiv preprint arXiv:1903.11027}, 2019.

\bibitem{funnelalgorithm}
J.~Hershberger and J.~Snoeyink, ``Computing minimum length paths of a given
  homotopy class,'' \emph{Computational geometry}, vol.~4, no.~2, pp. 63--97,
  1994.

\end{thebibliography}

\end{document}